\documentclass[11pt,a4paper]{article}
\usepackage{times,latexsym}
\usepackage{url}
\usepackage[T1]{fontenc}

\usepackage[acceptedWithA]{tacl2018v2}

\RequirePackage{amsmath} 
\usepackage{microtype} 
\usepackage{graphicx} 
\usepackage{bm} 
\usepackage{amsfonts} 
\usepackage[normalem]{ulem} 

\usepackage{multirow}
\usepackage{multicol}
\usepackage{array}
\usepackage{enumitem}
\makeatletter
\newif\if@restonecol
\makeatother

\usepackage[linesnumbered,ruled,vlined]{algorithm2e}
\usepackage{algpseudocode}

\usepackage{tabularx,booktabs}
\newcolumntype{Y}{>{\centering\arraybackslash}X} 

\makeatletter
\newcommand{\thickhline}{%
    \noalign {\ifnum 0=`}\fi \hrule height 1pt
    \futurelet \reserved@a \@xhline
}
\newcolumntype{"}{@{\hskip\tabcolsep\vrule width 1pt\hskip\tabcolsep}}
\makeatother

\usepackage{xspace,mfirstuc,tabulary}

\newif\iftaclinstructions
\taclinstructionsfalse 
\iftaclinstructions
\renewcommand{\confidential}{}
\renewcommand{\anonsubtext}{(No author info supplied here, for consistency with
TACL-submission anonymization requirements)}
\newcommand{\instr}
\fi

\iftaclpubformat 

\else

\fi

\title{A Neural Generative Model for Joint Learning Topics and Topic-Specific Word Embeddings}

\author{
 Lixing Zhu\textsuperscript{\dag} \qquad Yulan He\textsuperscript{\dag}\Thanks{Corresponding author.} \qquad Deyu Zhou\textsuperscript{\S} \\
 \textsuperscript{\dag}Department of Computer Science, University of Warwick, UK \\
 \textsuperscript{\S}School of Computer Science and Engineering, Key Laboratory of Computer Network\\
 and Information Integration, Ministry of Education, Southeast University, China \\
  {\sf \{lixing.zhu, yulan.he\}@warwick.ac.uk \qquad d.zhou@seu.edu.cn} \\
}

\date{}

\begin{document}
\maketitle
\begin{abstract}
We propose a novel generative model to explore both local and global context for joint learning topics and topic-specific word embeddings. In particular, we assume that global latent topics are shared across documents, a word is generated by a hidden semantic vector encoding its contextual semantic meaning, and its context words are generated conditional on both the hidden semantic vector and global latent topics. Topics are trained jointly with the word embeddings. The trained model maps words to topic-dependent embeddings, which naturally addresses the issue of word polysemy. Experimental results show that the proposed model outperforms the word-level embedding methods in both word similarity evaluation and word sense disambiguation. Furthermore, the model also extracts more coherent topics compared with existing neural topic models or other models for joint learning of topics and word embeddings. Finally, the model can be easily integrated with existing deep contextualized word embedding learning methods to further improve the performance of downstream tasks such as sentiment classification.
\end{abstract}

\section{Introduction}

Probabilistic topic models assume words are generated from latent topics which can be inferred from word co-occurrence patterns taking a document as global context. In recent years, various neural topic models have been proposed. Some of them are built on the Variational Auto-Encoder (VAE) \cite{kingma2014auto} which utilizes deep neural networks to approximate the intractable posterior distribution of observed words given latent topics \cite{miao2016neural,srivastava2017autoencoding,BouchacourtTN18}. However, these models take the bag-of-words (BOWs) representation of a given document as the input to the VAE and aim to learn hidden topics that can be used to reconstruct the original document. They do not learn word embeddings concurrently.

Other topic modeling approaches explore the pre-trained word embeddings for the extraction of more semantically coherent topics since word embeddings capture syntactic and semantic regularities by encoding the local context of word co-occurrence patterns. For example, the topic-word generation process in the traditional topic models can be replaced by generating word embeddings given latent topics \cite{das2015gaussian} or by a two-component mixture of a Dirichlet multinomial component and a word embedding component \cite{nguyen2015improving}. Alternatively, the information derived from word embeddings can be used to promote semantically-related words in the Polya Urn sampling process of topic models \cite{li2017enhancing} or generate topic hierarchies \cite{zhao2018inter}. However, all these models use pre-trained word embeddings and do not learn word embeddings jointly with topics.

While word embeddings could improve the topic modeling results, but conversely, the topic information could also benefit word embedding learning. Early word embedding learning methods~\cite{Mikolov13a} learn a mapping function to project a word to a single vector in an embedding space. Such one-to-one mapping cannot deal with word polysemy, as a word could have multiple meanings depending on its context. For example, the word `\emph{patient}' has two possible meanings `\emph{enduring trying circumstances with even temper}' and `\emph{a person who requires medical care}'. When analyzing reviews about restaurants and health services, the semantic meaning of `\emph{patient}' could be inferred depending on which topic it is associated with. One  solution is to first extract topics using the standard Latent Dirichlet Allocation (LDA) model and then incorporate the topical information into word embedding learning by treating each topic as a pseudo-word \cite{liu2015topical}. 

Whereas the aforementioned approaches adopt a two-step process, by either using pre-trained word embeddings to improve the topic extraction results in topic modeling, or incorporating topics extracted using a standard topic model into word embedding learning, \citet{shi2017jointly} developed a Skip-Gram based model to jointly learn topics and word embeddings based on the Probabilistic Latent Semantic Analysis (PLSA), where each word is associated with two matrices rather than a vector to induce topic-dependent embeddings. This is a rather cumbersome setup. \citet{foulds2018mixed} used the Skip-Gram to imitate the probabilistic topic model that each word is represented as an importance vector over topics for context generation.

In this paper we propose a neural generative model built on VAE, called the Joint Topic Word-embedding (JTW) model, for jointly learning topics and topic-specific word embeddings. More concretely, we introduce topics as tangible parameters that are shared across all the context windows. We assume that the pivot word is generated by the hidden semantics encoding the local context where it occurred. Then the hidden semantics is transformed to a topical distribution taking into account the global topics, and this enables the generation of context words. Our rationale is that the context words are generated by the hidden semantics of the pivot word together with a global topic matrix, which captures the notion that the word has multiple meanings that should be shared across the corpus. We are thus able to learn topics and generate topic-dependent word embeddings jointly. The results of our model also allow the visualization of word semantics because topics can be visualized via the top words and words can be encoded as distributions over the topics\footnote{Our source code is made available at \url{http://github.com/somethingx02/topical\_wordvec\_models}.}.

In summary, our contribution is three-fold:
\begin{itemize}[noitemsep] 
    \item We propose a novel Joint Topic Word-embedding (JTW) model built on VAE, for jointly learning topics and topic-specific word embeddings;
    \item We perform extensive experiments and show that JTW outperforms other Skip-Grams or Bayesian alternatives in both word similarity evaluation and word sense disambiguation tasks, and can extract semantically more coherent topics from data;
    \item We also show that JTW can be easily integrated with existing deep contextualized word embedding learning model to further improve the performance of downstream tasks such as sentiment classification.
\end{itemize}

\section{Related Work}

Our work is related to two lines of research:

\noindent\textbf{Skip-Gram approaches for word embedding learning}. 
The Skip-Gram, also known as \textsc{Word2Vec}~\cite{Mikolov13b}, maximizes the probability of the context words $\mathbf{w}_n$ given a centroid word $x_n$. \citet{pennington2014glove} pointed out that Skip-Gram neglects the global word co-occurrence statistics. They thus formulated the Skip-Gram as a non-negative matrix factorization (NMF) with the cross-entropy loss switched to the least square error. Another NMF-based method was proposed by~\citet{NIPS2018_7443}, in which the Euclidean distance was substituted with Wasserstein distance. \citet{jameel-schockaert-2019-word} rewrote the NMF objective as a cumulative product of normal distributions, in which each factor is multiplied by a von Mises-Fisher (vMF) distribution of context word vectors, to hopefully cluster the context words since the vMF density retains the cosine similarity. 

Although the Skip-Gram-based methods attracted extensive attention, they were criticized for their inability to capture the polysemy~\cite{pilehvar2016conflated}. A pioneered solution to this problem is the Multiple-Sense Skip-Gram (MSSG) model~\cite{neelakantan2014efficient}, where word vectors in a context are first averaged then clustered with other contexts to obtain a sense representation for the pivot word. In the same vein, \citet{iacobacci-navigli-2019-lstmembed} leveraged sense tags annotated by BabelNet~\cite{navigli2012babelnet} to jointly learn word and sense representations in the Skip-Gram manner that the context words are parameterized via a shared look-up table and sent to a BiLSTM to match the pivot word vector.

There have also been Bayesian extensions of the Skip-Gram models for word embedding learning. \citet{barkan2017bayesian} inherited the probabilistic generative line while extending the Skip-Gram by placing a Gaussian prior on the parameterized word vectors. The parameters were estimated via variational inference. In a similar vein, \citet{rios2018deep} proposed to generate words in bilingual parallel sentences by shared hidden semantics. They introduced a latent index variable to align the hidden semantics of a word in the source language to its equivalence in the target language. More recently, \citet{bravzinskas2018embedding} proposed the Bayesian Skip-Gram (BSG) model, in which each word type with its related word senses collapsed is associated with a `prior' or static embedding and then, depending on the context, the representation of each word is updated by `posterior' or dynamic embedding. Through Bayesian modeling, BSG is able to learn context-dependent word embeddings. It does not explicitly model topics, however. In our proposed JTW, global topics are shared among all documents and learned from data. Also, whereas BSG only models the generation of context words given a pivot word, JTW explicitly models the generation of both the pivot word and the context words with different generative routes. 

\noindent\textbf{Combining word embeddings with topic modeling}. 
Pre-trained word embeddings can be used to improve the topic modeling performance. For example, \citet{das2015gaussian} proposed the Gaussian LDA model, which, instead of generating discrete word tokens given latent topics, generates draws from a multivariate Gaussian of word embeddings. \citet{nguyen2015improving} also replaced the topic-word Dirichlet multinomial component in traditional topic models, but by a two-component mixture of a Dirichlet multinomial component and a word embedding component. \citet{li2017enhancing} proposed to modify the Polya Urn sampling process of the LDA model by promoting semantically-related words obtained from word embeddings. More recently, \citet{zhao2018inter} proposed to adapt a multi-layer Gamma Belief Network to generate topic hierarchies and also fine-grained interpretation of local topics, both of which are informed by word embeddings. 

Instead of using word embeddings for topic modeling, \citet{liu2015topical} proposed the Topical Word Embedding model which incorporates the topical information derived from standard topic models into word embedding learning by treating each topic as a pseudo-word. \citet{briakou-etal-2019-cross} followed this route and proposed a four-stage model in which topics were first extracted from a corpus by LDA and then the topic-based word embeddings are mapped to a shared space using anchor words which were retrieved from the WordNet. 

There are also approaches proposed to jointly learn topics and word embeddings built on Skip-Gram models. \citet{shi2017jointly} developed a Skip-Gram Topical word Embedding (STE) model built on PLSA where each word is associated with two matrices---one matrix used when the word is a pivot word and another used when the word is considered as a context word. Expectation-Maximization (EM) is used to estimate model parameters. \citet{foulds2018mixed} proposed the Mixed-Membership Skip-Gram model (MMSG), which assumes a topic is drawn for each context and the word in the context is drawn from the log-bilinear model based on the topic embeddings. Foulds trained their model by alternating between Gibbs sampling and noise-contrastive estimation. MMSG only models the generation of context words, but not pivot words.

While our proposed JTW also resembles the similarity to the Skip-Gram model in that it predicts the context word given the pivot word, it is different from the existing approaches in that it assumes global latent topics shared across all documents and the generation of the pivot word and the context words follows different generative routes. Moreover, it is built on VAE and is trained using neural networks for more efficient parameter inference.

\section{Joint Topic Word-embedding (JTW) Model}

\begin{figure*}[htb]
\centering
\includegraphics[width=1.0\textwidth]{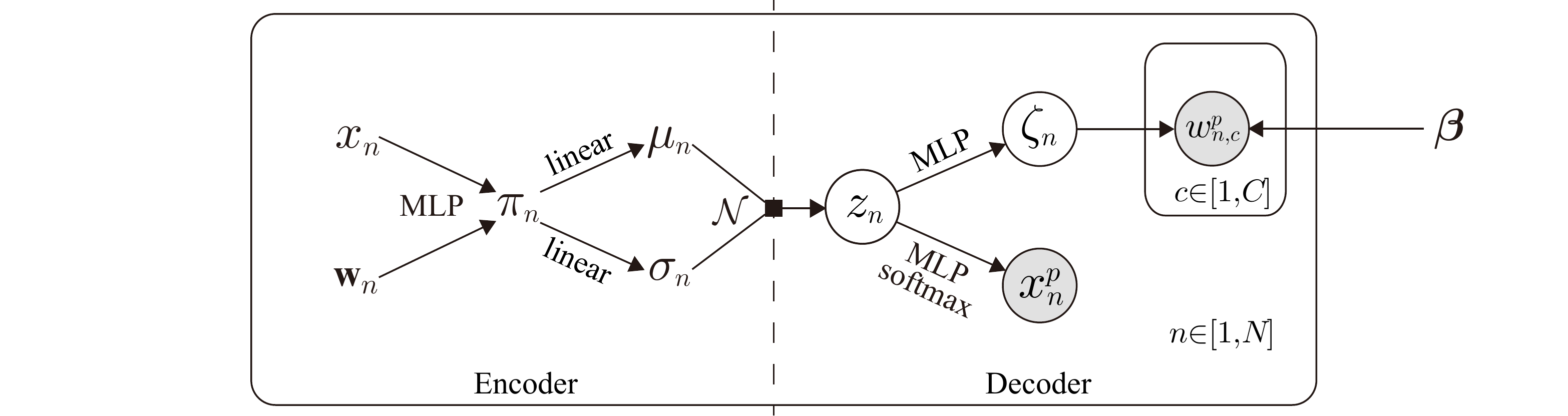}
\caption{The Variational Auto-Encoder framework for the Joint Topic Word-embedding (JTW) model. Boxes are ``plates'' indicating replicates. Shaded circles represent the observed variables. $\bm{\beta}$ is a $\mathrm{T}\times \mathrm{V}$ matrix representing corpus-wide latent topics.}\label{fig:2}
\end{figure*}

In this section, we describe our proposed Joint Topic Word-embedding (JTW) model built on VAE, as shown in Fig. \ref{fig:2}. We first give an overview of JTW, then present each component of the model, followed by the training details.

Following the problem setup in the Skip-Gram model, we consider a pivot word $x_n$ and its context window $\mathbf{w}_n=w_{n,1:C}$. We assume there are a total of $N$ pivot word tokens and each context window contains $C$ context words. However, as opposed to Skip-Gram, we do not compute the joint probability as a product chain of conditional probabilities of the context word given the pivot. Instead, in our model, context words are represented as BOWs for each context window by assuming the exchangeability of context words within the local context window.

We hypothesize that the hidden semantic vector $z_n$ of each word $x_n$ induces a topical distribution that is combined with the global corpus-wide latent topics to generate context words. Topics are represented as a probability matrix where each row is a multinomial distribution measuring the importance of each word within a topic. The hidden semantics $z_n$ of the pivot word $x_n$ is transformed to a topical distribution $\zeta_n$, which participates in the generation of context words. Our assumption is that each word embodies a finite set of meanings that can be interpreted as topics, thus each word representation can be transformed to a distribution over topics. Context words are generated by first selecting a topic and then sampled according to the corresponding multinomial distribution. This enables a quick understanding of word semantics through the topical distribution and at the same time learning the latent topics from the corpus. The generative process is given below:
\begin{itemize}
\item For each word position $n\in \{1,2,3,\dots,N\}$:
\begin{itemize}
\item Draw hidden semantic representation $z_n\sim\mathcal{N}(\bm{0},\bm{I})$
\item Choose a pivot word $x_n\sim p(x_n|z_n)$
\item Transform $z_n$ to $\zeta_n$ with a multi-layered perceptron: 
	$\zeta_n = \mathrm{MLP}(z_n)$
\item For each context word position $c\in \{1,2,3,\dots,C\}$:
\begin{itemize}
\item Choose a topic indicator $t_{n,c} \sim \mathrm{Categorical}(\zeta_n)$
\item Choose a context word $w_{n,c}\sim p(w_{n,c}|\beta_{t_{n,c}})$
\end{itemize}
\end{itemize}
\end{itemize}
Here, all the distributions are functions approximated by neural networks, e.g., $p(x_n|z_n) \propto \mathrm{exp}(\bm{M}_x z_n+\bm{b}_x)$, which will be discussed in more details in the Decoder section, $t_{n,c}$ indexes a row $\beta_{t_{n,c}}$ in the topic matrix. 
We could implicitly marginalise out the topic indicators, in which case the probability of a word would be written as $w_{n,c}|\zeta_n, \bm{\beta}\sim \mathrm{Categorical}(\sigma(\bm{\beta}^{\mathrm{T}} \zeta_n))$, where $\sigma(\cdot)$ denotes the softmax function. The prior distribution for $z_n$ is a multivariate Gaussian distribution with the mean $\bm{0}$ and covariance $\bm{I}$, of which the posterior indicates the hidden semantics of the pivot word when conditioned on $\{x_n,\mathbf{w}_n\}$.

Although both JTW and BSG assume that a word can have multiple senses and use a latent embedding $z$ to represent the hidden semantic meaning of each pivot word, there are some key differences in their generative processes. JTW first draws a latent embedding $z$ from a standard Gaussian prior which is deterministically transformed into topic distributions and a distribution over pivot words. The pivot word is conditionally independent of its context given the latent embedding. At the same time, each context word is assigned a latent topic, drawn from a shared topic distribution which leverages the global topic information, and then drawn independently of one another. In BSG the latent embedding $z$ is also drawn from a Gaussian prior but the context words are generated directly from the latent embedding $z$, as opposed to via a mixture model as in JTW. Therefore, JTW is able to group semantically-similar words into topics, which is not the case in BSG. 

Given the observed variables $\{x_{1:N},\mathbf{w}_{1:N}\}$, the objective of the model is to infer the posterior $p(\mathbf{z}|\mathbf{x},\mathbf{w})$. This is achieved by the VAE framework. As illustrated in Figure~\ref{fig:2}, the JTW model is composed of an encoder and a decoder, each of which is constructed by neural networks. The family of distributions to approximate the posterior is Gaussian, in which $\mu_n$ and $\sigma_n$ are optimized. As in VAE, we optimize $\mu_n$ and $\sigma_n$ through the training of parameters in neural networks (e.g., we optimize $\bm{M}_\pi$ in $\mu_n=\bm{M}_{\pi}^{\mathrm{T}}\pi_n+\bm{b}_{\pi}$ instead of updating $\mu_n$ directly).

\subsection{ELBO}
The VAE naturally simulates the variational inference~\cite{jordan1999introduction}, where a family of parameterized distributions $q_\phi(z_n|x_n, \mathbf{w}_n)$ are optimized to approximate the intractable true posterior $p_\theta(z_n|x_n, \mathbf{w}_n)$. This is achieved by minimizing the Kullback-Leibler (KL) divergence between the variational distribution and the true posterior for each data point:

\begin{equation}\label{eqs:1st}
\begin{aligned}
\mathrm{KL}(q_{\phi}(z_n|{x_n},\mathbf{w}_n)&||p_{\theta}(z_n|{x_n},\mathbf{w}_n))\\ 
=\mathrm{log}\,p_{\theta}(x_n, \mathbf{w}_n)-&\mathbb{E}_{q_{\phi}}\lbrack\mathrm{log}\,p_{\theta}(z_n,x_n,\mathbf{w}_n) \\
 &\ \ -\mathrm{log}\,q_{\phi}(z_n|x_n, \mathbf{w}_n)\rbrack,
\end{aligned}
\end{equation}
 \noindent where the expectation term is called the Evidence Lower Bound (ELBO), denoted as $\mathcal{L}(\theta,\phi;x_n,\mathbf{w}_n)$. VAE optimizes ELBO to presumably minimize the KL-divergence. The ELBO is further derived as 

\begin{equation}\label{eqs:4th}
\begin{aligned}
&\mathcal{L}(\theta,\phi;x_n, \mathbf{w}_n)\\
&=\mathbb{E}_{q_{\phi}(z_n|x_n,\mathbf{w}_n)}\left[\mathrm{log}\,p_{\theta}(x_n,\mathbf{w}_n|z_n)\right]\\
&\quad-\mathrm{KL}(q_{\phi}(z_n|x_n,\mathbf{w}_n)||p(z_n)).
\end{aligned}
\end{equation}
The first term on the left-hand side of Equation~\ref{eqs:4th}, which is an expectation with respect to $q_{\phi}(z_n|x_n,\mathbf{w}_n)$, can be estimated by sampling due to its intractability. That is:
\begin{equation}\label{eqs:6th}
\begin{aligned}
\mathbb{E}_{q_{\phi}(z_n|x_n,\mathbf{w}_n)}\left[\mathrm{log}\,p_{\theta}(x_n,\mathbf{w}_n|z_n)\right]\\
\approx \frac{1}{S}\sum_{s=1}^{S}\mathrm{log}\,p_{\theta}(x_n,\mathbf{w}_n|z_n^{(s)}),
\end{aligned}
\end{equation}
\noindent where $z_n^{(s)}\sim q_{\phi}(z_n|x_n,\mathbf{w}_n)$. Here we use $z_n^{(s)}$ to represent the samples since the sampled distribution is related to $x_n$.

\subsection{Encoder}
The Encoder corresponds to $q_{\phi}(z_n|{x_n},\mathbf{w}_n)$ in Equation~\ref{eqs:6th}. Recall that the variational family for approximating the true posterior is Gaussian Distribution parameterized by $\{\mu_n,\sigma_n\}$. As such, the encoder is essentially a set of neural functions mapping from observations to Gaussian parameters $\{\mu_n,\sigma_n\}$. The neural functions are defined as:
$\pi_n=\mathrm{MLP}(x_n, \mathbf{w}_n)$,
$\mu_n=\bm{M}^{\mathrm{T}}_\mu \pi_n + \bm{b}_\pi$,
$\sigma_n=\bm{M}^{\mathrm{T}}_\sigma \pi_n + \bm{b}_\sigma$,
where the MLP denotes the multi-layered perceptron and the context window $\mathbf{w}_n$ is represented as a BOW that is a $V$-dimentional vector. The encoder outputs Gaussian parameters $\{\mu_n,\sigma_n\}$, which constitutes the variational distribution $q_\phi(z_n|x_n, \mathbf{w}_n)$. In order to differentiate $q_\phi(z_n|x_n, \mathbf{w}_n)$ with respect to $\phi$, we apply the reparameterization trick~\cite{kingma2014auto} by using the
following transformation:
\begin{equation}\label{eqs:7th}
\begin{aligned}
z_n^{(s)}&=\mu_n + \sigma_n \odot \epsilon_n^{(s)}\\
&\epsilon_n^{(s)}\sim \mathcal{N}(\bm{0},\bm{I}).
\end{aligned}
\end{equation}

\subsection{Decoder}

The Decoder corresponds to $p_{\theta}(x_n,\mathbf{w}_n|z_n^{(s)})$ in Equation~\ref{eqs:6th}. It is a neural function that maps the sample $z_n^{(s)}$ to the distribution $p_{\theta}(x^p_n,\mathbf{w}^p_n|z_n^{(s)})$ with random variables instantiated by $x_n$ and $\mathbf{w}_n$. More concretely, we define two neural functions to generate the pivot word and the context words separately. Both the functions involve an MLP, while the context words are generated independently from each other by the topic mixture weighted by the hidden topic distributions. The neural functions are expressed as:
\begin{gather}
p(x^p_n|z_n^{(s)}) \propto \mathrm{exp}(\bm{M}_x z_n^{(s)}+\bm{b}_x)\label{eqs:4_r2}\\
\zeta_n^{(s)} = \mathrm{MLP}(z_n^{(s)}) \label{eqs:6_r2}\\
p(w^p_{n,c}|\zeta_n^{(s)}) \propto \mathrm{exp}(\bm{\beta}^{\mathrm{T}} \zeta_n^{(s)}+\bm{b}_w)\label{eqs:7_3}
\end{gather}

In this case, the MLP for the pivot word is specified as a fully-connected layer. Recall that we represent the context window $\mathbf{w}_n$ as BOW, the instantiated probability $p_{\theta}(x_n,\mathbf{w}_n|z_n^{(s)})$ can be therefore derived as:
\begin{equation}
\begin{aligned}\label{eqs:8th}
p_\theta(x_n,\mathbf{w}_{n}|z_n^{(s)}) \propto \exp(\bm{M}_x z_n^{(s)}+\bm{b}_x)\lbrack x_n\rbrack \\
\prod\nolimits_{v=1}^{V}\mathrm{exp}(\bm{\beta}^{\mathrm{T}} \zeta_n^{(s)}+\bm{b}_w)\lbrack v\rbrack^{\mathbf{w}_n\lbrack v\rbrack}
\end{aligned}
\end{equation}
where $\mathrm{exp}(\bm{M}_x z_n^{(s)}+\bm{b}_x)\lbrack x_n\rbrack$ denotes the $x_n$-th element of the vector $\mathrm{exp}(\bm{M}_x z_n^{(s)}+\bm{b}_x)$.
\subsection{Loss Function} 
We are now ready to compute ELBO in Equation~\ref{eqs:4th} with the specified $q_{\phi}(z_n|x_n,\mathbf{w}_n)$ and $p_\theta(x_n,\mathbf{w}_{n}|z_n^{(s)})$ in hand. 
Our final objective function that needs to be maximized is:
\begin{equation}\label{eqs:10th}
\begin{aligned}
&\mathcal{L}(\theta,\phi;x_n, \mathbf{w}_n) \\
&= \frac{1}{S}\sum_{s=1}^{S}\mathrm{log}\,p_{\theta}(x_n,\mathbf{w}_n|\mu_n +\sigma_n \odot \epsilon_n^{(s)})\\
&\quad-\frac{1}{2}\sum_{d=1}^{D}\left(1+\mathrm{log}\sigma_n\lbrack d \rbrack^2-\mu_n\lbrack d\rbrack^2-\sigma_n\lbrack d\rbrack^2\right),
\end{aligned}
\end{equation}
where $D$ denotes the dimension of $\mu$. $S$ denotes the number of sample points required for the computation of the expectation term. The loss function is the negative of the objective function.
The learning procedure is summarized in Algorithm~\ref{alg:1st}.

\setcounter{algocf}{0}
\begin{algorithm}
    \caption{Training of JTW model}
    \label{alg:1st}

    \KwIn{pivot words $x_{1:N}$, context windows $\mathbf{w}_{1:N}$, learning rate $\eta$, learning rate decay $lrDecay$, maximum iterative number $maxIter$, batch size $B$, batch number $N_B$\;}
    \KwOut{learned network parameters $\theta$,$\phi$\;}
    Initialize $\theta,\phi$ randomly\;
    $i \gets 0$, $\eta \gets 0.0005$\;
    For convenience, define $\mathbf{x}_{B} = x_{n:n+B}$, $\mathbf{w}_{B}=\mathbf{w}_{n:n+B}$ as a minibatch\; 
    \While{$\theta,\phi$ not converged \rm{and} $i < maxIter$}
    {
      Shuffle dataset $x_{1:N},\mathbf{w}_{1:N}$\;
      \For{$1$ \KwTo $N_B$}
      {
        Generate $S$ samples $\epsilon^{(s)}\sim \mathcal{N}(\bm{0},\bm{I})$\;
        Compute gradient $g\gets \nabla_{\theta,\phi}\mathcal{L}(\theta,\phi;\mathbf{x}_B, \mathbf{w}_B)$ according to Equation~\ref{eqs:10th}\;
        Update parameters $\theta,\phi$ using gradient $g$\;
      }
      $i \gets i+1$, $\eta \gets \eta \times lrDecay$\;
    }
    return $\theta,\phi$\;
\end{algorithm}

\subsection{Prediction} After training, we are able to map the words to their respective representations using the Encoder part of JTW. The Encoder takes a pivot word together with its context window as an input and outputs the parameters of the variational distribution considered to be the approximated posterior $q_\phi(z|x_n,\mathbf{w}_n)$, which is a Gaussian distribution in our case. The word representations are Gaussian parameters $\{\mu_n,\sigma_n\}$. Because the output of the Encoder is  formulated as a Gaussian distribution, the word similarity of two words can be either computed by the KL-divergence between the Gaussian distributions, or by the cosine similarity between their means. We use the Gaussian mean $\mu$ to represent a word given its context. The universal representation of a word type can be obtained by averaging the posterior means of all occurrences over the corpus.

\section{Experimental Setup}
\label{sec:expSetup}

\noindent\textbf{Dataset}. We train the proposed JTW model on the Yelp dataset\footnote{\url{https://www.yelp.com/dataset/documentation/main}}, which is a collection of more than 4 million 
reviews on over 140k 
business categories.
Although the number of business categories is large, the vast majority of reviews falls into $5$ business categories. The top \emph{Restaurant} category consists of more than 40\% of reviews. The next top 4 categories, \emph{Shopping}, \emph{Beauty\&Spas}, \emph{Automotive} and \emph{Clinical} contains about $8\%, 6\%, 4\%$ and $3\%$ of reviews, respectively. The \emph{Clinical} documents are further filtered by business subcategories defined in~\citet{tran2017online}, which are recognized as core clinical businesses. This results in $176,733$ documents for the \emph{Clinical} category. Because the dataset is extremely imbalanced, simply training the model on the original dataset will likely overfit to the \emph{Restaurant} category. We thus balance the dataset by sampling roughly an equal number of documents from each of the top 5 categories. The vocabulary size is set to $8,000$. We use Mallet\footnote{\url{http://mallet.cs.umass.edu/}} to filter out stopwords. The final dataset consists of $865,616$ documents with a total of $101,468,071$ tokens.

\noindent\textbf{Parameter Setting}. The word semantics are represented as $100$-dimensional vectors (i.e., $D=100$), which is a default configuration for word representations~\cite{Mikolov13a,bravzinskas2018embedding}. The number of latent topics is set to 50. It has been previously studied in~\citet{kingma2014auto} that the number of samples per data point can be set to $1$ if the batch size is large, (e.g. $>100$). In our experiments, we set the batch size to $2,048$ and the number of samples per data point, $S$, to $1$. The context window size is set to $10$. Network parameters (i.e., $\theta$, $\phi$) are all initialized by a normal distribution with zero mean and $0.1$ variance.

\noindent\textbf{Baselines}. We compare our model against four baselines:

\begin{itemize}
\item \textbf{CvMF}~\cite{jameel-schockaert-2019-word}. CvMF can be viewed as an extension of GloVe that modifies the objective function by multiplying a mixture of vMFs, whose distance is measured by cosine similarity instead of euclidean distance. The mixture depicts the underlying semantics with which the words could be clustered.

\item \textbf{Bayesian Skip-Gram (BSG)}~\cite{bravzinskas2018embedding}. BSG\footnote{\url{https://github.com/ixlan/BSG}} is a probabilistic word-embedding method built on VAE as well, which achieved the state-of-art among other Bayesian word-embedding alternatives~\cite{vilnis2015word,barkan2017bayesian}. BSG infers the posterior or dynamic embedding given a pivot word and its observed context and is able to learn context-dependent word embeddings. 

\item \textbf{Skip-gram Topical word Embedding (STE)}~\cite{shi2017jointly}. STE adapted the commonly known Skip-Gram by associating each word with an input matrix and an output matrix and used the Expectation-Maximization (EM) method with the negative sampling for model parameter inference.  
For topic generation, they need to evaluate the probability of $p(w_{t+j} | z, w_t)$ for each topic $z$ and each skip-gram $<w_t; w_{t+j}>$, and represent each topic as the ranked list of bi-grams.

\item \textbf{Mixed Membership Skip-Gram (MMSG)}~\cite{foulds2018mixed}. MMSG leverages
mixed membership modeling in which words are assumed to be clustered into topics and the words in the context of a given pivot word are drawn from the log-bilinear model using the vector representations of the context-dependent topic. Model inference is performed using the Metropolis-Hastings-Walker algorithm with noise-contrastive estimation.

\end{itemize}
Among the aforementioned baselines, CvMF and BSG only generate word embeddings and do not model topics explicitly. Also, CvMF only maps each word to a single word embedding whereas BSG can output context-dependent word embeddings. Both STE and MMSG can learn topics and topic-dependent embeddings at the same time. However, in STE the topic dependence is stored in the lines of word matrices and the word representations themselves are context independent. In contrast, MMSG associates each word with a topic distribution; it could produce contextualized word embeddings by summing up topic vectors weighed by the posterior topic distribution given a context. 
We probe into different topic counts and find the best setting for methods with topics or mixtures. In all the baselines, the dimensionality of word embeddings is tuned and finally set to $100$.

\section{Experimental Results}

We compare JTW with baselines on both word similarity and word-sense disambiguation tasks for the learned word embeddings, and also present the topic coherence and qualitative evaluation results for the extracted topics. Furthermore, we show that JTW can be easily integrated with deep contextualized word embeddings to further improve the performance of downstream tasks such as sentiment classification.

\subsection{Word Similarity} The word similarity task \cite{finkelstein2001placing} has been widely adopted to measure the quality of word embeddings. In the word similarity task, a number of pair-wise words are given. Each pair of words should be assigned with a score that indicates their relatedness. The calculated scores are then compared with the golden scores by means of Spearman rank-order correlation coefficient. Because the word similarity task requires context-free word representations, we aggregate all the occurrences and obtain a universal vector for each word. The distance used for similarity scores is cosine similarity. For STE, we use AvgSimC following~\citet{shi2017jointly}. We further make a comparison with the results of the Skip-Gram (SG)  model\footnote{\url{https://code.google.com/archive/p/word2vec/}}, which maps each word token to a single point in an Euclidean space without considering different senses of words. All the approaches are evaluated on the $7$ commonly used benchmarking datasets. For JTW, we average the results over 10 runs and also report the standard deviations. 
\begin{table*}
\centering
\caption{Spearman rank correlation coefficient on $7$ benchmarks.}\label{tab:2}
\fontsize{9pt}{12pt}\selectfont
\begin{tabular}{l|cccccc}
\toprule
Benchmarks  & SG & CvMF & BSG & STE & MMSG & JTW (std. dev.)\\
\midrule
WS353-SIM & \textbf{0.610}&0.597&0.529 & 0.582&0.579& 0.598 (.014)\\
WS353-ALL & 0.571& \textbf{0.615} & 0.551 & 0.538&0.558&0.606 (.012)\\
MEN & 0.649 & 0.632 & \textbf{0.656} & 0.650 & 0.627 & 0.653 (.006)\\
SimLex-999 & 0.321&0.313 & 0.271 & 0.301&0.281&\textbf{0.344} (.005)\\
SCWS & 0.620&0.637& \textbf{0.652} & 0.622& 0.624& 0.640 (.010)\\
MTurk771 & 0.548&0.524& 0.555 & 0.554&\textbf{0.596}&0.546 (.010)\\
MTurk287 & 0.534&0.517& 0.572 & \textbf{0.641}&0.599&0.639 (.006)\\
\midrule
Average & 0.550&0.548& 0.541 & 0.555& 0.552& \textbf{0.575} (.004)\\
\bottomrule
\end{tabular}
\end{table*}

The results are reported in Table~\ref{tab:2}. It can be observed that among the baselines, BSG achieves the lowest score on average, followed by MMSG. Although JTW clearly beats all the other models on SimLex-999 only, it only performs slightly worse than the top model in 5 out of the remaining 6 benchmarks. Overall, JTW gives the superior results on average. A noticeable gap can be observed on the Stanford's Contextual Word Similarities (SCWS) dataset where JTW, MMSG and BSG give better results compared with SG, CvMF and STE. This can be explained by the fact that, in SCWS, golden scores are annotated together with the context. However, SG, CvMF and STE can only produce context-independent word vectors. The results show the clear benefit of learning contextualized word vectors. 
Among the topic-dependent word embeddings, JTW built on VAE appears to be more effective than the PLSA-based STE and the mixed membership model MMSG, achieving the best overall score when averaging the evaluation results across all the seven benchmarking datasets. The small standard deviation of JTW indicates that the performance is consistent across multiple runs. 

\subsection{Lexical Substitution}

While the word similarity tasks focus more on the general meaning of a word (since word pairs are presented without context), in this section, we turn to the lexical substitution task~\cite{yuret2007ku,thater2011word}, which was designed to evaluate the word-embedding learning methods regarding their ability to disambiguate word senses. The lexical substitution task can be described by the following scenario: Given a sentence and one of its member words, find the most related replacement from a list of candidate words. As stated in~\citet{thater2011word}, a good lexical substitution should not only capture the relatedness between the candidate word and the original word, but also imply the correctness with respect to the context.

Following~\citet{bravzinskas2018embedding}, we derive the setting from~\citet{melamud2015simple} to ensure a fair comparison between the context-free word embedding methods and the context-dependent ones. In details, for JTW and BSG, we capture the context of a given word using the BOW representation, and derive the representation of each candidate word taking account of the context. For CvMF and STE, the similarity score is computed using
\begin{equation}
\mathrm{BalAdd}(x,y)=\frac{C\,\mathrm{cos}(y,x)+\sum_{c=1}^C\mathrm{cos}(y,w_c)}{2C},
\end{equation}
where $y$ is the candidate word and $x$ denotes the original word. For MMSG, the original word's representation is calculated as the sum of its associated topic vectors weighed by the word's posterior topical distribution. Given an original word and its context, we choose the candidate word with the highest similarity score. We compare the performance of various models on lexical substitution using the dataset from the SemEval 2007 task 10\footnote{\url{http://www.dianamccarthy.co.uk/task10index.html}}~\cite{mccarthy2007semeval}, which consists of 1,688 instances. Because some words have multiple synonyms as annotated in the dataset, we would consider a chosen candidate word as a correct prediction if it hits one of the ground-truth replacements.

\begin{table}
\centering
\fontsize{9pt}{12pt}\selectfont
\caption{Accuracy on the lexical substitution task.}\label{tab:3}
\begin{tabular}{l|ccccc}
\toprule
\textit{Model} & CvMF & BSG & STE & MMSG &  JTW\\
\midrule
\textit{Accuracy} &0.440& 0.453 & 0.433&0.474& \textbf{0.487}\\
\bottomrule
\end{tabular}
\end{table}

We report in Table~\ref{tab:3} the accuracy scores of different methods. Context-sensitive word embeddings generally perform better than context-free alternatives. STE can only learn context-independent word embeddings and hence gives the lowest score. BSG is able to learn context-dependent word embeddings and outperforms CvMF. Among the joint topic and word embedding learning methods, STE performs the worst, showing that associating each word with two matrices and learning topic-dependent word embddings based on PLSA appear to be less effective. Both JTW and MMSG show superior performances compared to BSG. JTW outperforms MMSG because JTW also models the generation of pivot word in addition to context words and the VAE framework for parameter inference is more effective than the annealed negative contrastive estimation used in MMSG.

\subsection{Topic Coherence}

\begin{figure}[hbp]
\raggedleft
\includegraphics[width=0.47\textwidth]{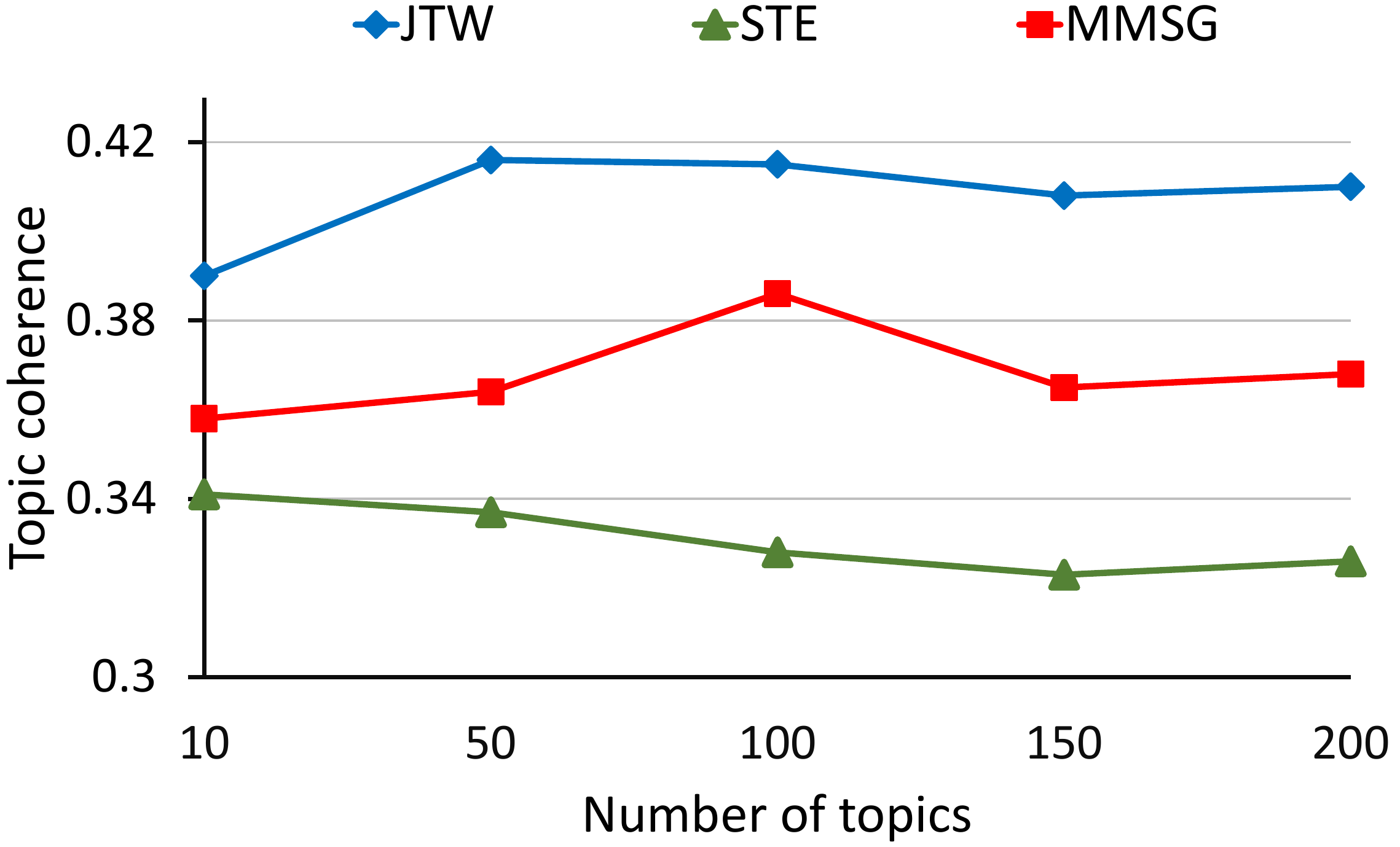}
\caption{Topic coherence scores versus number of topics.}\label{fig:2d2}
\end{figure}

Because only STE and MMSG can jointly learn topics and word embeddings among the baselines, we compare our proposed JTW with these two models in term of topic quality. The evaluation metric we employed is the topic coherence  metric proposed in~\citet{TopicEval2015}. The metric extracts co-occurrence counts of the topic words in Wikipedia using a sliding window of size 110. For each top word a vector is calculated whose elements are the normalized point-wise
mutual information between the word and
every other top words. Given a topic, the arithmetic mean of all vector pairs’ cosine similarity is treated as the coherence measure. We calculate the topic coherence score of each extracted topic based on its associated top ten words using Palmetto\footnote{\url{https://github.com/dice-group/Palmetto}}~\cite{Rosner2014}. The topic coherence results with the topic number varying between 10 and 200 are plotted in Figure~\ref{fig:2d2}. The graph shows that JTW scores the highest under all the topic settings. It gives the best coherence score of $0.416$ at $50$ topics, and gradually flattens with the increasing number of topics. 
MMSG exhibits an upward trend up to $100$ topics, and drops to $0.365$ when the topic number is set to $150$. STE undergoes a gradual decrease and then stabilizes with the topic number beyond $150$. 


\subsection{Extracted Topics}
\begin{table}[t]
\centering
\fontsize{8pt}{10pt}\selectfont
\caption{Example topics discovered by JTW and MMSG, each topic is represented by the top 10 words sorted by their likelihoods. The topic labels are assigned manually. Semantically less coherent words are highlighted by \emph{italics}.}\label{tab:4}
\begin{tabular}{ * {4}{c} c}
\toprule
\textbf{Topic 1} & \textbf{Topic 2} & \textbf{Topic 3} & \textbf{Topic 4} & \textbf{Topic 5}\\
\midrule
\emph{Food} & \emph{Shopping} & \emph{Beauty} & \makebox[\dimexpr(\width-0.5em)][r]{\emph{Automotive}} & \emph{Clinical}\\
\midrule
\multicolumn{5}{c}{\bf JTW}\\
\midrule
 good& great & hair & car & \makebox[\dimexpr(\width-1.4em)][r]{compassionate} \\ 
 food& friendly & \makebox[\dimexpr(\width-0.3em)][r]{recommend} & \emph{told} & caring \\
 chicken& service & highly & phone & personable \\
 place& staff & place & called & courteous\\
 pizza& shop & \makebox[\dimexpr(\width-0.4em)][r]{experience} & care & therapy\\
 love&  clean & fabulous & vehicle & competent\\
 cheese& helpful &  great & \makebox[\dimexpr(\width-0.4em)][r]{\emph{time}} & \makebox[\dimexpr(\width-1.4em)][r]{knowledgeable} \\ 
 salad& nice & nail &  BMW & passionate \\
 red& amazing & nails & insurance & physician\\
 delicious& customer &  awesome & wanted & respectful\\
\midrule
\multicolumn{5}{c}{\bf MMSG}\\
\midrule
food & friendly &  massage &  place & therapy\\
service & staff & spa &  service & physical\\
great & great & back & \emph{time} & pain \\
good & helpful & great & \emph{back} & \emph{back}\\
place & service & \emph{time} & customer & \emph{massage} \\
\emph{friendly} & clean & good & car & \makebox[\dimexpr(\width-1.0em)][r]{recommend} \\
\emph{staff} & place & massages & \emph{people} & \makebox[\dimexpr(\width-0.4em)][r]{great}\\
nice & nice & facial & good & therapist \\
\emph{back} & store & \emph{therapist} & money & \emph{work} \\
prices & super &  body & \emph{give} & highly\\
\bottomrule
\end{tabular}
\end{table}
We present in Table~\ref{tab:4} the example topics extracted by JTW and MMSG. It can be easily inferred from the top words generated by JTW that Topic 1 is related to `\emph{Food}', whereas Topic 5 is about the `\emph{Clinical Service}', which is identified by the words `\emph{caring}' and `\emph{physician}'. It can also be deduced from the top words that Topic 2, 3 and 4 represent `\emph{Shopping}', `\emph{Beauty}' and `\emph{Automotive}', respectively. In contrast, topics produced by MMSG contain more semantically less coherent words as highlighted by italics. For example, Topic 1 in MMSG contains words relating to both food and staff. This might be caused by the fact that, in MMSG, training is performed as a two-stage process by first assigning topics to words using Gibbs Sampling then estimating the topic vectors and word vectors from word co-occurrences and topic assignments via maximum likelihood estimator. This is equivalent to a topic model with parameterized word embeddings. Conversely, in JTW, latent variables in the generative process are recognized as word representations. Parameters reside in the generative network, and are inferred by the VAE. No extra parameters are introduced to encode the words. Therefore, the topics extracted tend to be more identifiable.

\subsection{Visualization of Word Semantics}

\begin{figure}[htb]
\centering
\includegraphics[width=0.49\textwidth]{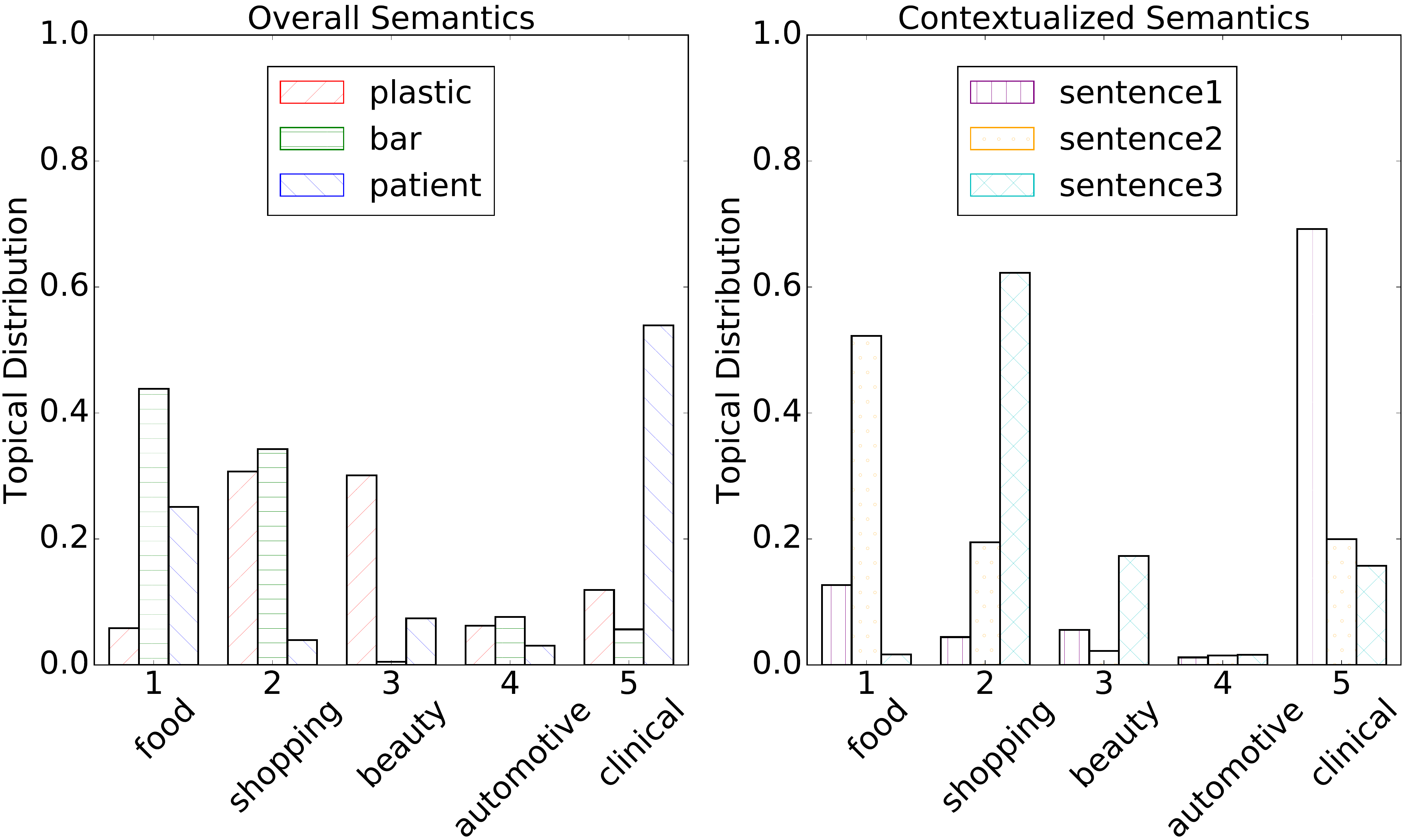}
\caption{The overall topical distributions and contextualized topical distributions of the example words and the contextualized topical distribution of three example sentences. Note that the $x$-axis denotes the five example topics shown in Table 4.}\label{fig:3}
\end{figure}

The extracted topics allow the visualization of word semantics. In JTW, a word's semantic meanings can be interpreted as a distribution over the discovered latent topics. This is achieved by aggregating all the contextualized topical distribution of a particular word throughout the corpus. Meanwhile, when a word is placed under a specific context, its topical distribution can be directly transformed from its contextualized representation. We chose three words---`\emph{plastic}', `\emph{bar}' and `\emph{patient}'---to illustrate the polysemous nature of them. To further demonstrate their context-dependent meanings, we also visualize the topic distribution of the following three sentences: (1) \emph{Effective patient care requires clinical knowledge and understanding of physical therapy}; (2) \emph{Restaurant servers require patient temperament}; (3) \emph{You have to bring your own bags or boxes but you can also purchase plastic bags}. The topical distribution for the pivot words and the three example sentences are shown in Figure~\ref{fig:3}.

We can deduce from the overall distributions that the semantic meaning of `\emph{plastic}' distributes almost equally on two topics, `\emph{shopping}' and `\emph{beauty}', while the meaning of `\emph{bar}' is more prominent on the `\emph{food}' and `\emph{shopping}' topics. `\emph{Patient}' has a strong connection with the `\emph{clinical}' topic, though it is also associated with the `\emph{food}' topic. When considering a specific context about the patient care, Sentence 1 has its topic distribution peaked at the `\emph{clinical}' topic. Sentence 2 also contains the word `\emph{patient}', but it now has its topic distribution peaked at `\emph{food}'. Sentence 3 mentioned `\emph{plastic bags}' and its most prominent topic is `\emph{shopping}'. These results show that JTW can indeed jointly learn latent topics and topic-specific word embeddings.

\begin{table*}
\centering
\caption{Results on the 5-class sentiment classification by 10-fold cross validation on the Yelp reviews.}\label{tab:6}
\begin{tabular}{l|l|l|l|l}
\thickhline
\multicolumn{1}{c|}{\multirow{2}{*}{Model}} & \multicolumn{4}{c}{Criteria}\\
\cline{2-5}
&\multicolumn{1}{c|}{Precision} &\multicolumn{1}{c|}{Recall} & \multicolumn{1}{c|}{Macro-F1} & \multicolumn{1}{c}{Micro-F1}\\
\hline
JTW & 0.5713$\pm$.021 & 0.5639$\pm$.014 & 0.5599$\pm$.016 & 0.7339$\pm$.015\\
ELMo & 0.6091$\pm$.005 & 0.6053$\pm$.001 & 0.6056$\pm$.002 & 0.7610$\pm$.005 \\
BERT & 0.6293$\pm$.014 & 0.5952$\pm$.006 & 0.6041$\pm$.012 & 0.7626$\pm$.005\\
JTW-ELMo & 0.6286$\pm$.008 & \textbf{0.6110}$\pm$\textbf{.004} & \textbf{0.6168}$\pm$\textbf{.008} & 0.7783$\pm$.004\\
JTW-BERT & \textbf{0.6354$\pm$\textbf{.014}} & 0.6081$\pm$.009& 0.6045$\pm$.014 &\textbf{0.7806}$\pm$\textbf{.005}\\
\thickhline
\end{tabular}
\end{table*}

\subsection{Integration with Deep Contextualized Word Embeddings}

Recent advances in deep contextualized word representation learning have generated significant impact in natural language processing. Different from traditional word embedding learning methods such as Word2Vec or GloVe, where each word is mapped to a single vector representation, deep contextualized word representation learning methods are typically trained by language modelling and generate a different word vector for each word depending on the context in which it is used. A notable work is ELMo~\cite{peters2018deep}, which is commonly regarded as the pioneer for deriving deep contextualized word embeddings~\cite{devlin2018bert}. 
ELMo calculates the weighed sum of different layers of a multi-layered BiLSTM-based language model, using the normalized vector as a representation for the corresponding word. More recently, in contrast to ELMo, BERT \cite{devlin2018bert} was proposed to apply the bidirectional training of Transformer to masked language modelling. Because of its capability of effectively encoding contextualized knowledge from huge external corpora in word embeddings, BERT has refreshed the state-of-art results on a number of NLP tasks.

While Word2Vec/GloVe and ELMo/BERT represent the two opposite extremes in word embedding learning, with the former learning a single vector representation for each word and the latter learning a separate vector representation for each occurrence of a word, our proposed JTW sits in the middle that it learns different word vectors depending on which topic a word is associated with. Nevertheless, we can incorporate ELMo/BERT embeddings into JTW. This is achieved by replacing the BOW input with the pre-trained ELMo/BERT word embeddings in the Encoder-Decoder architecture of JTW, making the resulting word embeddings better at capturing semantic topics in a specific domain. More precisely, the training objective is switched to the cosine value of half the angle between the input ELMo/BERT vector and decoded output vector formulated as:
\begin{equation}\label{eqs:11th}
\begin{aligned}
p_\theta(x_n,\mathbf{w}_{n}|z_n^{(s)}) \propto  \mathrm{cos}(\frac{1}{2}\mathrm{arccos}(\frac{x^{\top}_n \cdot x^{(p)}_n}{\Vert x_n\Vert \Vert x_n^{(p)} \Vert}))\\
\prod\nolimits_{c=1}^{C}\mathrm{cos}(\frac{1}{2}\mathrm{arccos}(\frac{w^{\top}_{n,c} \cdot w^{(p)}_{n,c}}{\Vert w_{n,c}\Vert \Vert w_{n,c}^{(p)} \Vert})),
\end{aligned}
\end{equation}
where $x_n^{(p)}$ and $w_{n,c}^{(p)}$ are the reconstructed representations generated from $z_n^{(s)}$ by Equation~\ref{eqs:4_r2} and Equation~\ref{eqs:7_3}, respectively. Recall that, the input to the model has been encoded by pre-trained word vectors (e.g., 300-dimensional vectors). Our training objective is to make the reconstructed $x_n^{(p)}$ and $w_{n,c}^{(p)}$ as close as possible to their original input word embeddings. The difference is measured by the angle between the input and the output vectors. Normalized ELMo/BERT vectors can be transformed to the polar coordinate system with trigonometric functions, which forms a probability distribution by 
\begin{equation}
\int\nolimits_{0}^{\pi}\frac{1}{2}\mathrm{cos}\frac{\theta}{2}\mathrm{d}\theta=1,
\end{equation}
and the function is monotone to the similarity between the input ELMo/BERT embeddings and the reconstructed output embeddings, which reaches its peak when $x_n = x_n^{(p)}$ (i.e., $\theta=0$). Therefore, we are able to replace Equation~\ref{eqs:8th} with Equation~\ref{eqs:11th} when an ELMo/BERT is attached. The input vectors of the Encoder are then the embeddings produced by ELMo/BERT, and the Decoder output are the reconstructed word embeddings aligned with the input. 

We resort to the sentiment classification task on Yelp and compare the performance of JTW, ELMo and BERT\footnote{\url{https://github.com/google-research/bert}}, and the integration of both, JTW-ELMo and JTW-BERT, by 10-fold cross validation. In all the experiments, we fine-tune the models on the training set consisting of 90\% documents sampled from the dataset described in Section~\ref{sec:expSetup} and evaluate on the $10\%$ data that serves as the test set. We employ the further pre-training scheme \cite{sun2019fine} that different learning rates are applied to each layer and slanted triangular learning rates are imposed across epochs when adapting the language model to the training corpus \cite{howard2018universal}. The classifier used for all the methods is an attention hop over a BiLSTM with a softmax layer. The ground truth labels are the five-scale review ratings included in the original dataset. The 5-class sentiment classification results in precision, recall, macro-F1 and micro-F1 scores are reported in Table~\ref{tab:6}.

It can be observed from Table~\ref{tab:6} that a sentiment classifier trained on JTW-produced word embeddings gives worse results compared with that using the deep contextualized word embeddings generated by ELMo or BERT. Nevertheless, when integrating the ELMo or BERT front-end with JTW, the combined model, JTW-ELMo and JTW-BERT, outperforms the original deep contextualized word representation models, respectively. It has been verified by the paired $t$-test that JTW-ELMo outperforms ELMo and BERT at the $95\%$ significance level on Micro-F1. The results show that our proposed JTW is flexible and it can be easily integrated with pre-trained contextualized word embeddings to capture the domain-specific semantics better compared to directly fine-tuning the pre-trained ELMo or BERT on the target domain, hence leading to improved sentiment classification performance.

\section{Conclusion}
Driven by the motivation that combining word embedding learning and topic modeling can mutually benefit each other, we propose a probabilistic generative framework that can jointly discover more semantically coherent latent topics from the global context and also learn topic-specific word embeddings, which naturally address the problem of word polysemy. Experimental results verify the effectiveness of the model on word similarity evaluation and word sense disambiguation. Furthermore, the model can discover latent topics shared across documents, and the encoder of JTW can generate the topical distribution for each word. This enables an intuitive understanding of word semantics. We have also shown that our proposed JTW can be easily integrated with deep contextualized word embeddings to further improve the performance of downstream tasks. In future work, we will explore the discourse relationships between context windows to model, for example, the semantic shift between the neighboring sentences.

\section*{Acknowledgments}
The authors would like to thank the anonymous reviewers for insightful comments and helpful suggestions. This work was funded in part by EPSRC (grant no. EP/T017112/1). LZ was funded by the Chancellor's International Scholarship at the University of Warwick. DZ was partially funded by the National Key Research and Development Program of China (2017YFB1002801) and the National Natural Science Foundation of China (61772132).

\bibliography{tacl2018}
\bibliographystyle{acl_natbib}

\end{document}